\documentclass{article}

\def\redc{\cellcolor[HTML]{FF999A}}
\def\orangec{\cellcolor[HTML]{FFCC99}}
\def\yellowc{\cellcolor[HTML]{FFF8AD}}
\usepackage[table]{xcolor}
\definecolor{lightyellow}{RGB}{255, 255, 224}
\newcolumntype{:}{!{\vrule width 0.8pt}}
\newcommand{\ours}{TATRA}

\usepackage{microtype}
\usepackage{graphicx}
\usepackage{adjustbox}
\usepackage{booktabs,array,colortbl,xcolor,makecell}
\usepackage{subcaption}
\usepackage{booktabs} 
\usepackage{placeins}

\usepackage{algorithm}
\usepackage{algorithmic}

\usepackage{enumitem}
\setlist[itemize]{
    leftmargin=15pt,      
    itemsep=0pt,          
    parsep=0pt,           
    topsep=0pt,        
    partopsep=0pt      
}

\usepackage{tikz}
\usetikzlibrary{positioning, fit, calc, shapes.geometric, backgrounds, shadows.blur, arrows.meta, decorations.pathreplacing, decorations.pathmorphing, shapes.symbols, shapes.callouts}

\definecolor{headerbg}{HTML}{0EA5A2} 
\definecolor{headerfg}{HTML}{FFFFFF} 
\definecolor{rowalt}{HTML}{F8FAFC}   
\definecolor{yescol}{HTML}{16A34A}   
\definecolor{nocol}{HTML}{DC2626}   
\definecolor{maybe}{HTML}{F59E0B}   
\definecolor{pillbg}{HTML}{E5E7EB}   %
\definecolor{coreBlue}{RGB}{66, 133, 244}
\definecolor{coreRed}{RGB}{234, 67, 53}
\definecolor{coreGreen}{RGB}{52, 168, 83}
\definecolor{coreYellow}{RGB}{251, 188, 5}
\definecolor{slate}{RGB}{70, 80, 90}

\usepackage{graphicx}
\usepackage[most]{tcolorbox}
\tcbuselibrary{breakable,listings}
\usepackage{longtable}
\usepackage{booktabs}
\usepackage{array}
\newcolumntype{C}{c}

\newtcblisting{promptbox}[2][]{%
  breakable,
  colback=white,
  colframe=teal!70!black,
  boxrule=0.8pt,
  arc=2mm,
  left=2mm,right=2mm,top=1.5mm,bottom=1.5mm,
  fonttitle=\bfseries,
  title=#2,
  listing only,
  listing options={
    basicstyle=\ttfamily\small,
    breaklines=true,
    breakatwhitespace=false,
    columns=fullflexible
  },
  #1
}

\newcommand{\Badge}[1]{%
  \tikz[baseline=-0.6ex, scale=0.22]{
    \draw[line width=0.9pt, draw=black!12, fill=black!2] (0,0) circle (1.9);
    #1
  }%
}

\newcommand{\YesMark}{%
\scalebox{0.5}{
  \Badge{%
    \draw[line width=2.2pt, draw=yescol, line cap=round, line join=round]
      (-1.0,-0.2) -- (-0.2,-1.0) -- (1.1,0.9);
  }%
  }
}

\newcommand{\NoMark}{%
\scalebox{0.5}{
  \Badge{%
    \draw[line width=2.0pt, draw=nocol, line cap=round] (-1.1,-1.1) -- (1.1,1.1);
    \draw[line width=2.0pt, draw=nocol, line cap=round] (-1.1,1.1) -- (1.1,-1.1);
  }%
  }
}

\newcommand{\WaveMark}{%
\scalebox{0.5}{
  \Badge{%
    \draw[
      line width=2.0pt, draw=maybe,
      decorate, decoration={snake, amplitude=0.55pt, segment length=3.2pt}
    ] (-1.3,0) -- (1.3,0);
  }%
  }
}

\usepackage{hyperref}

\usepackage[preprint]{icml2026}
\usepackage[utf8]{inputenc}
\usepackage[T1]{fontenc}

\usepackage{amsmath}
\usepackage{amssymb}
\usepackage{mathtools}
\usepackage{amsthm}

\usepackage[capitalize,noabbrev]{cleveref}

\theoremstyle{plain}

\theoremstyle{definition}

\theoremstyle{remark}

\usepackage[textsize=tiny]{todonotes}

\icmltitlerunning{Training-Free Instance-Adaptive Prompting Through Rephrasing and Aggregation}

\begin{document}

\twocolumn[
  \icmltitle{\ours{}: \underline{T}raining-Free Instance-\underline{A}daptive Prompting \underline{T}hrough \underline{R}ephrasing and \underline{A}ggregation}

  \icmlsetsymbol{equal}{*}

  \begin{icmlauthorlist}
   \icmlauthor{Bartosz Dziuba}{equal,yyy}
    \icmlauthor{Kacper Kuchta}{equal,yyy}
    \icmlauthor{Pawe\l{} Batorski}{equal,comp}
    \icmlauthor{Przemys\l{}aw Spurek}{yyy,sch}
    \icmlauthor{Paul Swoboda}{comp}
  \end{icmlauthorlist}
  

  \icmlaffiliation{yyy}{Jagiellonian University}
  \icmlaffiliation{comp}{Heinrich Heine Universität Düsseldorf}
  \icmlaffiliation{sch}{IDEAS Research Institute}

  \icmlcorrespondingauthor{}{bartosz.dziuba@student.uj.edu.pl}

  \icmlkeywords{Machine Learning, ICML}

  \vskip 0.3in

\scalebox{0.25}{ \includegraphics{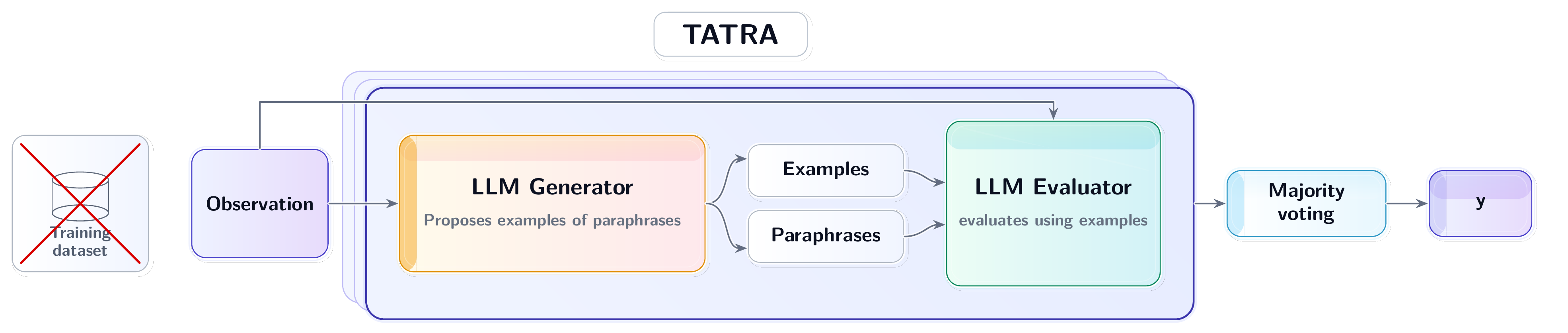} }

\scalebox{0.25}{ \includegraphics{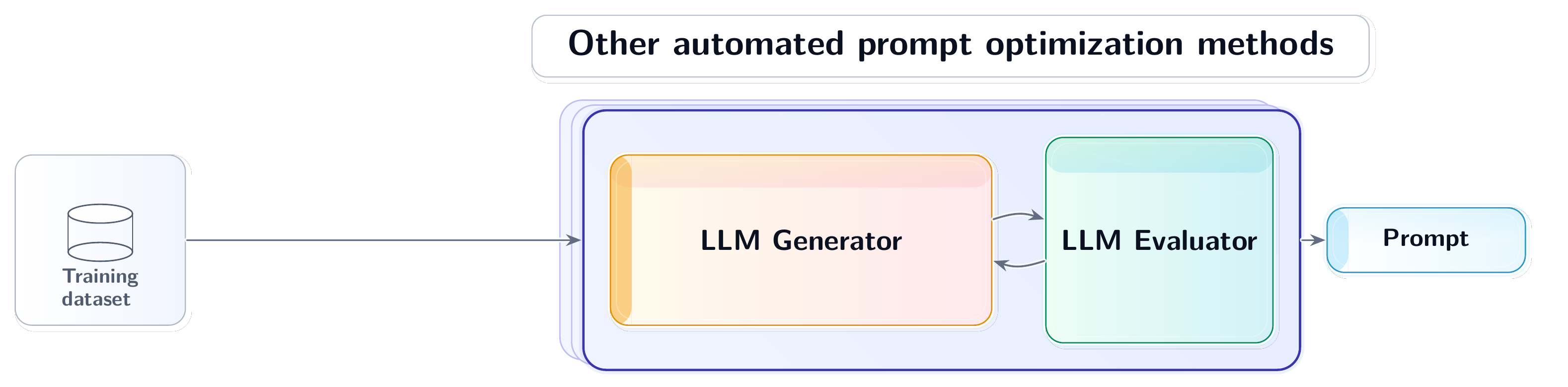} }
\captionof{figure}{Comparison of \ours{} to existing automated prompt-engineering methods. Most prior approaches require a task-specific training set and run expensive dataset-level optimization loops to produce a single prompt per task. In contrast, \ours{} is training-free and dataset-free, constructing a small set of instance-specific few-shot demonstrations on the fly and aggregating predictions across rephrasings for robust per-sample prompting.}

]

\printAffiliationsAndNotice{\icmlEqualContribution}

\begin{abstract}
Large Language Models (LLMs) have improved substantially alignment, yet their behavior remains highly sensitive to prompt phrasing. 
This brittleness has motivated automated prompt engineering, but most existing methods (i) require a task-specific training set, (ii) rely on expensive iterative optimization to produce a single dataset-level prompt, and (iii) must be rerun from scratch for each new task. We introduce \ours{}, a dataset-free prompting method that constructs instance-specific few-shot prompts by synthesizing on-the-fly examples to accompany a user-provided instruction. \ours{} requires no labeled training data and avoids task-specific optimization loops, while retaining the benefits of demonstration-based prompting. Across standard text classification benchmarks, \ours{} matches or improves over strong prompt-optimization baselines that depend on training data and extensive search. On mathematical reasoning benchmarks, \ours{} achieves state-of-the-art performance on GSM8K and DeepMath, outperforming methods that explicitly optimize prompts on those tasks. Our results suggest that per-instance construction of effective in-context examples is more important than running long, expensive optimization loops to produce a single prompt per task.
We will make all code publicly available upon acceptance of the paper.
Code is available at \url{https://github.com/BMD223/TATRA}
\end{abstract}

\section{Introduction}
Large Language Models (LLMs) have advanced rapidly through scaling and post-training, including reinforcement learning from human feedback (RLHF) \citep{christiano2017deep, stiennon2020learning, ouyang2022training}, reinforcement learning with verifiable rewards (RLVR) for tasks with automatically checkable outcomes \citep{rlvr,shao2024deepseekmath,guo2025deepseek, math500}, and direct preference-based objectives such as DPO \citep{rafailov2023direct}; however, even highly capable aligned models remain brittle to prompt form, as small semantics-preserving changes in wording, formatting, or ordering can cause large performance swings \citep{sclar2023quantifying, zhuo2024prosa, errica2025did, razavi2025benchmarking, cao2024worst, chatterjee2024posix}, with robustness further degrading under human-plausible perturbations and prompt-injection-style edits \citep{zhu2023promptrobust, li2024evaluating, zhan2024injecagent}. This sensitivity echoes classic in-context learning effects, calibration, demonstration choice, and exemplar ordering \citep{zhao2021calibrate, lu2022fantastically, min2022rethinking}—and persists even for modern RL-aligned reasoning models \citep{guo2025deepseek, grace}, motivating adaptive prompting methods beyond a single static dataset-level prompt.

A natural response to prompt brittleness is Automated Prompt Engineering (APE) \citep{ape}, which replaces manual instruction design with algorithmic search over prompts; early work such as Automated Prompt Engineer \citep{ape} performs a one-shot generate-and-select procedure over candidate instructions, but it neither iteratively refines prompts, nor supports few-shot prompts with in-context demonstrations, and it relies on a task-specific dataset for evaluation and selection. Subsequent research has mainly addressed the first two limitations by introducing multi-iteration search-and-improve pipelines \citep{grace, evoprompt, batorski2025prl, piast, apo}; for example, APO \citep{apo} can incorporate few-shot demonstrations but largely by reusing training examples, which may be scarce or mismatched, while PRL and PIAST can synthesize or introduce additional demonstrations yet still assume access to a task-specific training set during prompt construction, leaving the dataset-free setting unresolved. 

In contrast, relatively few methods explicitly aim to eliminate the need for any task-specific dataset. A notable exception is GPS \citep{batorski2025gps}, which targets dataset-free prompt construction; however, it relies on substantial RLVR-style optimization and does not center on generating strong few-shot demonstrations for new tasks, which can limit downstream gains. More broadly, we argue that existing APE approaches remain constrained in practice: when many tasks arise ad hoc, repeatedly producing a high-quality prompt per task can still be costly and unwieldy, especially if the procedure implicitly requires significant compute or data-like supervision.

In this work, we introduce \ours{}, a general prompter that constructs per-instance prompts without requiring any task-specific training dataset. For each input, \ours{} synthesizes a small set of on-the-fly few-shot examples to pair with a user-provided instruction, combining dataset-free prompt generation with the empirical benefits of demonstration-based prompting. Concretely, \ours{} runs a lightweight iterative procedure: it generates multiple rephrasings of the input and candidate example sets, queries an LLM to label the original input and its rephrasings under these generated prompts, and repeats this process to obtain diverse, independent predictions. The final output is then produced by majority vote over all collected labels, and the entire pipeline is executed independently for each test instance.

In summary, our contributions are as follows:
\begin{itemize}
    \item We introduce \ours{}, a dataset-free prompting method that constructs instance-specific few-shot prompts by synthesizing in-context examples on the fly.
    
    \item We show that \ours{} matches or surpasses strong prompt-optimization baselines on standard text classification benchmarks without using task-specific training data, suggesting that per-instance few-shot example construction can be more effective than long, task-level prompt optimization loops.
    
    \item We demonstrate strong results on mathematical reasoning benchmarks, text classification and Domain-Based task.
\end{itemize}

\section{Related Work}

\paragraph{Prompt Engineering}
Prompting can substantially improve LLM performance without updating model parameters, making it a lightweight and cost-effective alternative to retraining~\citep{liu2023pretrain}. A large line of work studies how prompt design can better elicit reasoning by imposing structure on the model’s generation process. Chain-of-Thought (CoT) prompting encourages models to produce intermediate reasoning steps~\citep{wei2022chain}, while Tree-of-Thought (ToT) extends this idea by exploring multiple candidate reasoning trajectories~\citep{yao2023tree}. Other methods encode structure directly into the prompt: Program-of-Thoughts separates symbolic computation from natural-language reasoning~\citep{chen2022program}, and Graph-of-Thoughts organizes reasoning as a graph of interdependent steps~\citep{besta2024graph}. Least-to-Most prompting decomposes complex problems into simpler subproblems to improve compositional generalization~\citep{zhou2023leasttomost}. Complementary techniques such as zero-shot CoT and self-consistency aim to improve robustness by altering how reasoning traces are elicited and aggregated~\citep{kojima2022large,wang2022self}. In contrast to these primarily instruction- or structure-centric approaches, few-shot prompting conditions the model on a small set of in-context exemplars~\citep{brown2020language}, and has proven effective across diverse settings including puzzles and evidence extraction~\citep{xu2023llms,greenblatt2024getting,sivarajkumar2024empirical}.

\paragraph{Automated Prompt Engineering}
Automated prompt engineering (APE) seeks to replace manual prompt design with algorithmic procedures that search over instructions and, in some cases, demonstrations. Early work such as APE adopts a generate-and-rank pipeline: it produces candidate instructions and selects the one that performs best on the target task~\citep{ape}. Subsequent methods introduce iterative refinement. APO improves prompts via repeated natural-language critique and revision, although its few-shot demonstrations are drawn from the available training set~\citep{apo}. A broad family of approaches explores alternative optimization mechanisms: EvoPrompt applies evolutionary operators to evolve prompt candidates~\citep{evoprompt}, while PromptAgent frames prompt optimization as planning, using MCTS with error-driven feedback~\citep{wang2023promptagent}. Promptbreeder studies self-referential prompt evolution in which prompts propose and select improved variants of themselves~\citep{fernando2023promptbreeder}, and related work generalizes this perspective to open-ended, self-replicating optimization processes~\citep{wang2025evolving}. GRACE introduces gated refinement together with adaptive prompt compression~\citep{grace}, and OPRO treats the LLM as a black-box optimizer that proposes improved instructions given past candidates and their scores~\citep{yang2024llm_optimizers}. Complementary directions operate at the token level: AutoPrompt searches over discrete prompt tokens using gradient-based signals~\citep{shin2020autoprompt}, and~\citep{lu2024strings} show that even simple random sampling can serve as a surprisingly strong baseline.

\noindent
Another line of work applies reinforcement learning. RLPrompt learns short token prompts via RL~\citep{deng2022rlprompt}, while PRL can also synthesize in-context examples when beneficial~\citep{batorski2025prl}. Among the methods above, only APO and PRL explicitly incorporate examples: APO reuses training examples, whereas PRL can introduce novel examples but often requires tens of hours of compute, limiting practicality. PIAST aims to craft prompts with in-text examples efficiently, but still assumes access to a dataset and must be rerun per dataset~\citep{piast}.

In contrast, \ours{} generates per-sample few-shot examples that are not present in the training data, and does so without any task-specific training, eliminating the need for a labeled dataset. We summarize comparisons against the baselines we test against in Table~\ref{tab:comparison}.

\paragraph{Multi-agent / multi-module prompt optimization.}
Recent work optimizes prompts for multi-stage LM programs and agentic pipelines rather than a single LLM call. DSPy provides a framework and ``teleprompting'' optimizers~\citep{dspy}; MiPRO~\citep{mipro} and GePA~\citep{gepa} optimize instructions and demonstrations across modules; BetterTogether combines prompt optimization with parameter updates in modular pipelines~\citep{bettertogether}; and SGLang targets efficient execution of structured LM programs~\citep{sglang}. We do not compare to these methods because they assume a multi-module / multi-agent program and optimize end-to-end behavior, whereas \ours{} produces a single per-sample prompt for one LLM call.

\paragraph{Demonstration selection/retrieval.}
Demonstration selection constructs an input-specific prompt by retrieving in-context examples from a fixed pool, most commonly the labeled training set. \citet{rubin-etal-2022-learning} learn a dense retriever using LM-scored (input, demonstration) helpfulness labels; \citet{li-etal-2023-unified} train a retriever that generalizes across task families with a unified list-wise ranking objective; and Skill-KNN performs instance-wise selection using engineered representations~\citep{an-etal-2023-skill}. MoD further partitions the demonstration pool into expert subsets to support collaborative retrieval~\citep{wang2024mixture}. We do not compare to these approaches because they require per-instance retrieval and continued access to a demonstration bank at inference time, whereas \ours{} generates the prompt (including examples) on the fly for each new observation and does not depend on persistent access to a fixed pool of demonstrations.

\begin{table*}[ht]
\centering
\begingroup
\rowcolors{2}{rowalt}{white}
\caption{Qualitative comparison of prompt-engineering methods. We contrast manual prompting and automated prompt-engineering algorithms along four axes: Dataset-Free (does not require task-specific labeled data to construct the prompt), Training-Free (does not rely on gradient-based training or RL optimization for new prompt creation), Few-shot (supports prompts with in-context demonstrations), and Per-Sample (adapts the prompt to each input rather than producing a single dataset-level prompt). \ours{} is the only method in this comparison that is simultaneously dataset-free and training-free while also generating per-sample few-shot prompts.}

\begin{adjustbox}{max width=0.97\textwidth}
\begin{tabular}{@{} l c c c c @{}}
\rowcolor{headerbg}
\multicolumn{1}{@{}l}{\color{headerfg}\bfseries Algorithm} &
\multicolumn{1}{c}{\color{headerfg}\bfseries Dataset-Free} &
\multicolumn{1}{c}{\color{headerfg}\bfseries Training-Free} &
\multicolumn{1}{c}{\color{headerfg}\bfseries Few\mbox{-}shot} &
\multicolumn{1}{c@{}}{\color{headerfg}\bfseries Per-Sample} \\
\midrule

Manual Instruction~\citep{zhang2022opt} & \YesMark  & \YesMark   & \NoMark    & \NoMark  \\

APE~\citep{ape}              & \NoMark   & \YesMark   & \NoMark    & \NoMark  \\

APO~\citep{apo}       & \NoMark   & \YesMark   & \WaveMark & \NoMark  \\

EvoPrompt~\citep{evoprompt}      & \NoMark   & \YesMark    & \NoMark    & \NoMark  \\

PRL~\citep{batorski2025prl}            & \NoMark   & \NoMark    & \YesMark  & \NoMark  \\

GRACE~\citep{grace}                    & \NoMark   & \YesMark   & \NoMark    & \NoMark  \\

GPS~\citep{batorski2025gps}            & \YesMark  & \YesMark   & \NoMark    & \YesMark \\

\ours{}                                & \YesMark  & \YesMark   & \YesMark  & \YesMark \\
\bottomrule
\label{tab:comparison}
\end{tabular}
\end{adjustbox}
\endgroup

\vspace{4pt}
{\footnotesize
\textbf{Legend:}\;
\YesMark\ yes \;\;
\NoMark\ no \;\;
\WaveMark\ partial.
}

\end{table*}

\section{Method}
We present \ours{}, a training-free prompting pipeline that improves robustness to prompt phrasing without updating model parameters and without using any task-specific supervision.

\paragraph{Overview}
Our approach consists of five core components. 
(i)~First, we provide an instruction via a system prompt that outlines the task, the permissible outputs, and expected behavior.
(ii)~Second, we generate $k$ in-context examples balanced across task labels.
(iii)~Third, we paraphrase the test input $n$ times to safeguard results against linguistic variation.
(iv)~Fourth, we evaluate the original and paraphrased inputs using a frozen language model with constructed prompts.
(v) Fifth, we perform majority voting across all paraphrased variants.

Steps (i)--(iv) are repeated $r$ times, and the final prediction is obtained by voting over the total set of predictions across all runs. Importantly, we do not use any supervision in any way on the target datasets. All fixed hyperparameters (including the number of paraphrases $n$, in-context examples $k$, and repeated runs $r$) are reported in Appendix~\ref{sec:appendix_hyperparams}.

\begin{figure*}[]
\centering
\includegraphics[width=0.8\linewidth]{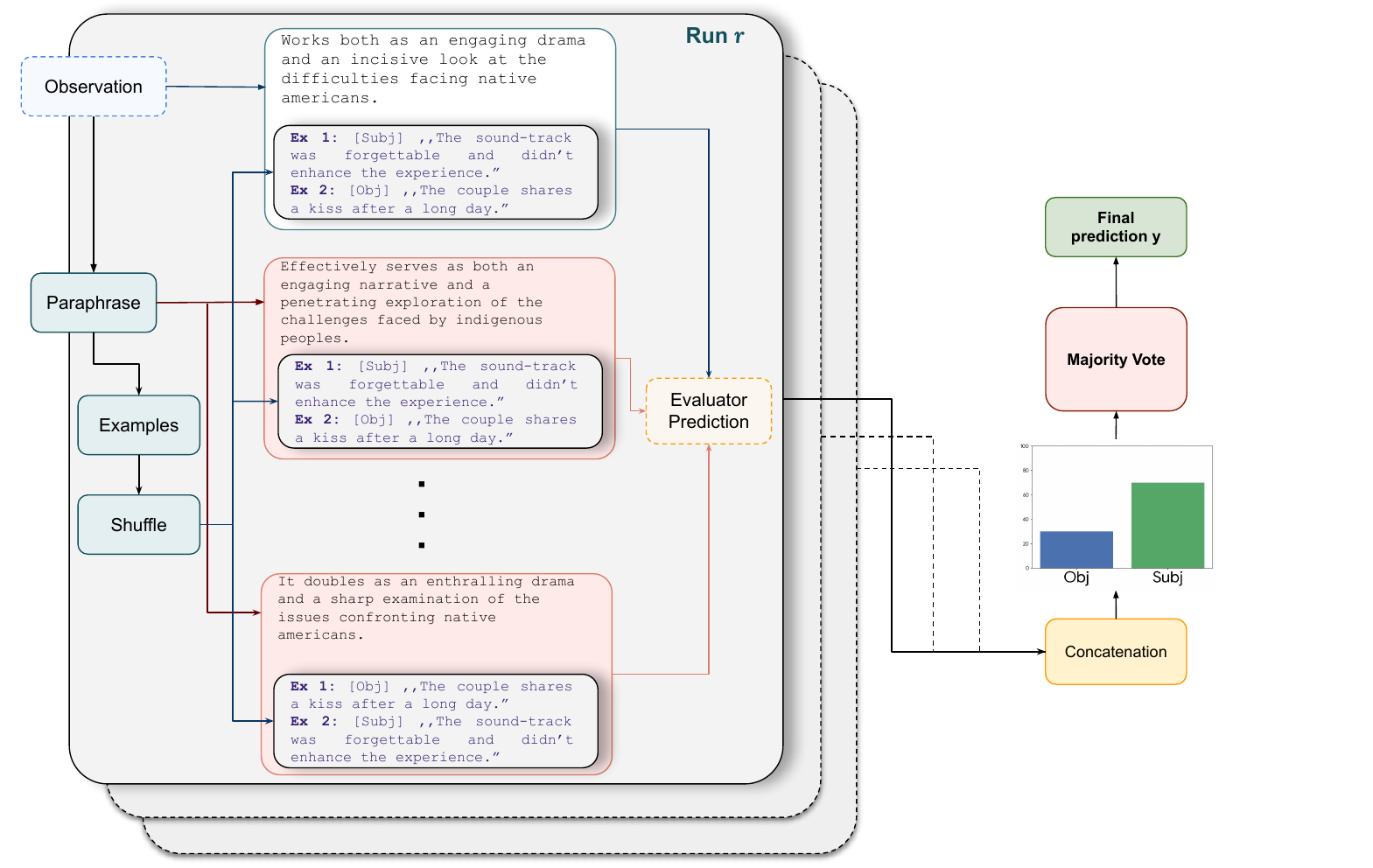}
\caption{
Overview of the TATRA prompting pipeline for subjective/objective classification. The process begins with an initial observation. A generator model produces multiple semantically equivalent paraphrases (e.g., swapping "native americans" for "indigenous peoples") to ensure linguistic robustness. Simultaneously, a set of label-balanced, synthetic few-shot examples is generated. These examples are shuffled and concatenated to both the original observation and its paraphrases.
An evaluator model then provides a prediction for each variant.
This paraphrase and in-context example generation is run $r$ times independently.
Finally, all individual predictions are aggregated via majority vote (showing a strong bias toward the "Subj" label in this instance) to produce the final prediction.
}
\label{fig:tatra_pipeline}
\end{figure*}


\paragraph{In-Context Example Generator}
Given a label set $\mathcal{Y}$, we generate for each label corresponding in-context examples. To get a balanced set of in-context examples, we sample a uniform number of examples for each label.

In detail, for each label $\ell \in \mathcal{Y}$, we ask the Example Generator LLM to generate in-context examples, specifying additionally:
\begin{itemize}
    \item The number of examples for each category.
    \item Strict formatting requirements, (e.g.\ for sentence classification the format should be \texttt{Sentence: <text> Label: $\ell$}).
    \item Style rules specific to the label $\ell$ (e.g.\ favorable adjectives for positive sentiment such as ``praise, satisfaction, recommendation'', critical ones for negative sentiment, such as ``criticism, disappointment, complaints'').
    \item The number of examples to generate.
    \item Fine-grained topic category (e.g.\ see Appendix~\ref{app:topicpools})
    \item Sentence-count constraints for generating a mix of short/medium/long in-context examples.
\end{itemize}

After generating in-context examples, we filter out in-context examples with errors, e.g.\ mismatched label. The in-context example generator produces a candidate set $\{e_1, e_2, \ldots, e_k\}$.




\paragraph{Input Prompt Paraphraser}
The paraphraser generates semantically equivalent variations of an input prompt while preserving task-critical information (e.g., label, topic, mathematical meaning).

For a given input prompt $x$ and target count $n$ of paraphrases we ask the Paraphrasing LLM to construct task-specific paraphrase prompts taking into account:
  \begin{itemize}
    \item the input prompt $x$.
    \item the number of paraphrases $n$ to generate,
    \item preservation constraints (e.g., ``keep tone and sentiment identical'' for sentiment tasks; ``preserve the mathematical problem and answer'' for math tasks),
    \item an output format requirement (``output only the paraphrases, one per line'').
  \end{itemize}


The paraphraser returns $n$ semantically equivalent variants $\{x_1^{(p)}, \ldots, x_n^{(p)}\}$ for each test input $x$. We then form the candidate set
\[
\mathcal{C}(x) = \{x\} \cup \{x_1^{(p)}, x_2^{(p)}, \ldots, x_n^{(p)}\},
\]
so that each instance is evaluated under $n{+}1$ paraphrases. All candidates in $\mathcal{C}(x)$ are scored using the same instruction and the same fixed set of synthesized in-context examples.

\paragraph{Prompt Evaluator}
The prompt evaluator is a frozen language model that classifies each candidate by concatenating three components: (i) the original or paraphrased test input, (ii) a prefix that specifies the task instructions and includes the fixed in-context examples, and (iii) a suffix that elicits the final label prediction.



\paragraph{Multi-Run Aggregation}
For each test instance, we obtain predictions $\hat{y}_{i,j}$ for the original input and its $n$ paraphrases across $r$ independent runs, where $i \in \{0,\ldots,n\}$ indexes the input variant (with $i=0$ denoting the original) and $j \in \{1,\ldots,r\}$ indexes the run. We then aggregate all $(n+1)\cdot r$ predictions with unweighted majority voting:
\begin{equation}
\hat{y} = \arg\max_{\ell \in \mathcal{Y}} \left| \left\{ (i,j) : \hat{y}_{i,j} = \ell \right\} \right| \,,
\end{equation}
where $|\cdot|$ denotes set cardinality. In the case of ties, we break ties deterministically by selecting the prediction obtained on the original, non-paraphrased input.


Pseudo-code for our method is given in Algorithm ~\ref{alg:pipeline} and a graphical illustration in Figure~\ref{fig:tatra_pipeline}.



\begin{algorithm}[t]
\caption{\ours{}}
\label{alg:pipeline}
\begin{small}
\begin{algorithmic}[1]
\STATE \textbf{Input:} Generator $\mathcal{M}_{\text{GEN}}$, Evaluator $\mathcal{M}_{\text{EVAL}}$
\STATE \textbf{Hyperparameters:} Examples $k$, Paraphrases $n$, Runs $r$
\STATE \textbf{Data:} Test set $\mathcal{D} = \{(x_i, y_i)\}_{i=1}^{N}$
\STATE \textbf{Output:} Predictions $\{\hat{y}_i\}_{i=1}^{N}$

\STATE \COMMENT{Phase 1: Example Generation (Once per run)}
\STATE $\mathcal{E} \gets \textsc{GenExamples}(\mathcal{M}_{\text{GEN}}, k)$
\STATE $\mathcal{E} \gets \textsc{Validate}(\mathcal{E})$
\STATE $P_{\text{ctx}} \gets \textsc{FormatPrefix}(\mathcal{E})$

\STATE \COMMENT{Phase 2: Evaluation Loop}
\FOR{$i \gets 1$ \textbf{to} $N$}
    \STATE $\mathcal{V} \gets \emptyset$ \COMMENT{Initialize set of all predicted labels for $x_i$}
    \FOR{$t \gets 1$ \textbf{to} $r$}
        \STATE \COMMENT{Generate $n$ paraphrases for input $x_i$}
        \STATE $\mathcal{C} \gets \{x_i\} \cup \textsc{Paraphrase}(\mathcal{M}_{\text{GEN}}, x_i, n)$
        
        \FOR{$x \in \mathcal{C}$}
            \STATE $prompt \gets \textsc{Concat}(P_{\text{ctx}}, x)$
            \STATE $\hat{y} \gets \mathcal{M}_{\text{EVAL}}(prompt)$
            \STATE $\mathcal{V} \gets \mathcal{V} \cup \{\hat{y}\}$ \COMMENT{Collect predictions from all variants and runs}
        \ENDFOR
    \ENDFOR
    
    \STATE \COMMENT{Majority vote on all concatenated runs}
    \STATE $\hat{y}_i \gets \textsc{MajorityVote}(\mathcal{V})$
\ENDFOR

\STATE \textbf{return} $\{\hat{y}_i\}_{i=1}^{N}$
\end{algorithmic}
\end{small}
\end{algorithm}

\section{Experiments}

\subsection{Experimental Setup}
All experiments in which \texttt{Qwen2.5-Instruct-7B} \citep{yang2024qwen2}, served as both the generator and the evaluator were conducted on a single NVIDIA A100 GPU. 
Scaling model experiments with larger models and cross models ablations were run on a single NVIDIA GH200 GPU. 
To ensure a fair comparison, all baseline methods reported in this paper were run using \texttt{Qwen2.5-Instruct-7B}, so that differences in performance do not arise from the choice of the underlying LLM. For cross-model studies, we have chosen \texttt{Llama-3.1-8B-Instruct}.

In this section, we compare \textit{TATRA} against literature baselines across three task families: classification, mathematical reasoning, and domain-based tasks. 
Unless stated otherwise, all results (including ablations) are averaged over three runs (three random seeds). We use a single set of fixed hyperparameters across all standard test runs (excluding hyperparameter exploration):
\begin{itemize}
    \item Paraphrases per item: $n = 10$
    \item In-context examples: $k = 16$
    \item Repeated runs: $r = 15$
\end{itemize}
Full details regarding hyperparameter configuration are provided in Appendix~\ref{sec:appendix_hyperparams}.

\subsection{Baselines}
\begin{itemize}
    \item \textbf{MI (Manual Instruction)}~\citep{zhang2022opt} \& \textbf{NI (Natural Instruction)}~\citep{mishra2021cross}:
    Human-written, task-specific prompts used as a manual prompting baseline.

    \item \textbf{APE (Automatic Prompt Engineer)}~\citep{ape}:
    Generates a pool of candidate instructions with an LLM and selects the best one via downstream performance; it does not add demonstrations (i.e., it performs instruction rephrasing only).

    \item \textbf{APO (Automatic Prompt Optimization)}~\citep{apo}:
    Treats prompt design as black-box optimization with iterative refinement, and can include few-shot demonstrations from the training set.

    \item \textbf{EvoPrompt}~\citep{evoprompt}:
    Evolves instruction prompts using evolutionary operators (selection/crossover/mutation); like APE, it focuses on instruction variants rather than generating new demonstrations.
    \begin{itemize}
        \item \textbf{DE (Differential Evolution)}: explores the prompt space via differential-evolution updates.
        \item \textbf{GA (Genetic Algorithm)}: improves prompts via standard genetic operators.
    \end{itemize}

    \item \textbf{GRACE}~\citep{grace}:
    Iteratively refines prompts with a no-regression acceptance rule and applies adaptive compression to control prompt length.

    \item \textbf{GPS}~\citep{batorski2025gps}:
    A dataset-free prompt construction approach. We report two variants:
    \begin{itemize}
        \item \textbf{GPS-J}: uses an LLM-as-a-judge regularizer.
        \item \textbf{GPS-SR-0.1}: applies sample regularization with probability $0.1$.
    \end{itemize}

    \item \textbf{PRL (Prompts from Reinforcement Learning)}~\citep{batorski2025prl}:
    Uses reinforcement learning to generate and optimize prompts, and can synthesize few-shot examples beyond the training set.

    \item \textbf{PIAST}~\citep{piast}:
    A prompt-and-example construction method evaluated under two compute/data budgets:
    \begin{itemize}
        \item \textbf{PIAST}: medium runtime budget with limited access to the training set.
        \item \textbf{PIAST (E)}: extended runtime budget with full dataset access.
    \end{itemize}
\end{itemize}

\subsection{Results}

\begin{table*}[ht]
\centering
\renewcommand{\arraystretch}{1.0}
\setlength{\tabcolsep}{3pt}
\caption{Accuracy on classification tasks, averaged over three runs.
Colours mark the best (\redc{red}), second-best (\orangec{orange}) and third-best (\yellowc{yellow}) numbers in each column; minor differences ($\le{}0.05$) are treated as ties.
The right-most column shows the mean accuracy of each method across the seven datasets.
Methods are grouped by whether prompt construction is dataset-free.}
\resizebox{0.895\textwidth}{!}{%
\begin{tabular}{lccccccc|c}
\multicolumn{9}{c}{\textbf{Not dataset-free (requires a task training set for prompt construction)}}\\
\hline
\textbf{Method} & \textbf{SST-2} & \textbf{CR} & \textbf{MR} & \textbf{SST-5} & \textbf{AG's News} & \textbf{TREC} & \textbf{Subj} & \textbf{Avg} \\
\hline
APO
& 93.71\textsubscript{\textcolor{gray}{$\pm$0.25}}
& \redc{93.48\textsubscript{\textcolor{gray}{$\pm$0.24}}}
& 89.97\textsubscript{\textcolor{gray}{$\pm$1.37}}
& 53.94\textsubscript{\textcolor{gray}{$\pm$0.29}}
& 83.73\textsubscript{\textcolor{gray}{$\pm$0.31}}
& 71.30\textsubscript{\textcolor{gray}{$\pm$1.90}}
& 69.80\textsubscript{\textcolor{gray}{$\pm$5.96}}
& 79.42 \\
APE
& 91.23\textsubscript{\textcolor{gray}{$\pm$0.66}}
& \yellowc{92.87\textsubscript{\textcolor{gray}{$\pm$0.02}}}
& 89.90\textsubscript{\textcolor{gray}{$\pm$0.94}}
& 49.37\textsubscript{\textcolor{gray}{$\pm$5.66}}
& 82.58\textsubscript{\textcolor{gray}{$\pm$1.20}}
& 77.07\textsubscript{\textcolor{gray}{$\pm$1.61}}
& 73.92\textsubscript{\textcolor{gray}{$\pm$1.39}}
& 79.56 \\
GA
& 94.65\textsubscript{\textcolor{gray}{$\pm$1.04}}
& 92.75\textsubscript{\textcolor{gray}{$\pm$0.40}}
& 90.45\textsubscript{\textcolor{gray}{$\pm$0.72}}
& 53.76\textsubscript{\textcolor{gray}{$\pm$1.13}}
& 82.24\textsubscript{\textcolor{gray}{$\pm$1.00}}
& \orangec{79.20\textsubscript{\textcolor{gray}{$\pm$2.83}}}
& 74.93\textsubscript{\textcolor{gray}{$\pm$3.12}}
& 81.14 \\
DE
& 93.29\textsubscript{\textcolor{gray}{$\pm$0.34}}
& \orangec{93.38\textsubscript{\textcolor{gray}{$\pm$0.19}}}
& 89.98\textsubscript{\textcolor{gray}{$\pm$0.24}}
& \orangec{55.25\textsubscript{\textcolor{gray}{$\pm$0.37}}}
& 82.18\textsubscript{\textcolor{gray}{$\pm$1.04}}
& 76.47\textsubscript{\textcolor{gray}{$\pm$0.38}}
& 73.08\textsubscript{\textcolor{gray}{$\pm$4.95}}
& 80.52 \\
GRACE
& 93.61\textsubscript{\textcolor{gray}{$\pm$0.53}}
& 90.92\textsubscript{\textcolor{gray}{$\pm$1.15}}
& 89.60\textsubscript{\textcolor{gray}{$\pm$1.51}}
& 53.96\textsubscript{\textcolor{gray}{$\pm$0.93}}
& 82.34\textsubscript{\textcolor{gray}{$\pm$0.39}}
& 72.53\textsubscript{\textcolor{gray}{$\pm$8.62}}
& 73.92\textsubscript{\textcolor{gray}{$\pm$3.05}}
& 79.55 \\
PRL
& \redc{96.32\textsubscript{\textcolor{gray}{$\pm$0.04}}}
& 92.83\textsubscript{\textcolor{gray}{$\pm$0.24}}
& \orangec{91.27\textsubscript{\textcolor{gray}{$\pm$0.05}}}
& \redc{56.21\textsubscript{\textcolor{gray}{$\pm$0.15}}}
& 84.36\textsubscript{\textcolor{gray}{$\pm$0.08}}
& 77.07\textsubscript{\textcolor{gray}{$\pm$2.36}}
& \yellowc{76.90\textsubscript{\textcolor{gray}{$\pm$0.95}}}
& \yellowc{82.14} \\
PIAST
& 95.35\textsubscript{\textcolor{gray}{$\pm$0.14}}
& 92.35\textsubscript{\textcolor{gray}{$\pm$0.05}}
& 90.57\textsubscript{\textcolor{gray}{$\pm$0.21}}
& 53.27\textsubscript{\textcolor{gray}{$\pm$0.66}}
& \orangec{85.93\textsubscript{\textcolor{gray}{$\pm$0.62}}}
& 77.07\textsubscript{\textcolor{gray}{$\pm$3.30}}
& 75.93\textsubscript{\textcolor{gray}{$\pm$0.40}}
& 81.50 \\
PIAST (E)
& \yellowc{95.88\textsubscript{\textcolor{gray}{$\pm$0.24}}}
& 92.55\textsubscript{\textcolor{gray}{$\pm$0.35}}
& \yellowc{91.00\textsubscript{\textcolor{gray}{$\pm$0.65}}}
& 53.33\textsubscript{\textcolor{gray}{$\pm$0.35}}
& \redc{87.39\textsubscript{\textcolor{gray}{$\pm$0.35}}}
& \yellowc{78.40\textsubscript{\textcolor{gray}{$\pm$1.22}}}
& \orangec{80.98\textsubscript{\textcolor{gray}{$\pm$0.67}}}
& \orangec{82.79} \\
\midrule

\multicolumn{9}{c}{\textbf{Dataset-free prompt construction}}\\
\hline
\textbf{Method} & \textbf{SST-2} & \textbf{CR} & \textbf{MR} & \textbf{SST-5} & \textbf{AG's News} & \textbf{TREC} & \textbf{Subj} & \textbf{Avg} \\
\hline
MI
& 92.70 & 87.25 & 87.40 & 52.31 & 82.29 & 69.20 & 57.95 & 75.59 \\
NI
& 95.77 & 91.50 & 90.85 & 51.90 & 83.43 & 66.60 & 68.10 & 78.31 \\
GPS-J
& 94.25\textsubscript{\textcolor{gray}{$\pm$1.20}}
& 90.65\textsubscript{\textcolor{gray}{$\pm$0.05}}
& 89.15\textsubscript{\textcolor{gray}{$\pm$0.38}}
& \yellowc{55.16\textsubscript{\textcolor{gray}{$\pm$0.36}}}
& 84.04\textsubscript{\textcolor{gray}{$\pm$0.02}}
& 72.80\textsubscript{\textcolor{gray}{$\pm$0.60}}
& 64.20\textsubscript{\textcolor{gray}{$\pm$2.25}}
& 78.61 \\
GPS-SR-0.1
& 92.98\textsubscript{\textcolor{gray}{$\pm$0.19}}
& 90.50\textsubscript{\textcolor{gray}{$\pm$0.38}}
& 88.70\textsubscript{\textcolor{gray}{$\pm$0.05}}
& 55.14\textsubscript{\textcolor{gray}{$\pm$1.13}}
& 84.21\textsubscript{\textcolor{gray}{$\pm$0.34}}
& 68.20\textsubscript{\textcolor{gray}{$\pm$0.20}}
& 65.10\textsubscript{\textcolor{gray}{$\pm$0.28}}
& 77.83 \\

\ours{}
& \orangec96.18\textsubscript{\textcolor{gray}{$\pm$0.31}}
& 92.75\textsubscript{\textcolor{gray}{$\pm$0.10}} 
& \redc91.78\textsubscript{\textcolor{gray}{$\pm$0.28}}
& 54.66\textsubscript{\textcolor{gray}{$\pm$0.41}}
& \yellowc85.61\textsubscript{\textcolor{gray}{$\pm$0.19}}
& \redc86.15\textsubscript{\textcolor{gray}{$\pm$0.61}}
& \redc82.18\textsubscript{\textcolor{gray}{$\pm$0.23}} 
& \redc84.19\\
\bottomrule

\end{tabular}%
}
\label{tab:cls}
\end{table*}

\paragraph{Classification.}
We evaluate our method on a suite of classification benchmarks spanning sentiment analysis, question classification, news categorization, and subjectivity detection:
\begin{itemize}
    \item \textbf{Binary sentiment classification:} SST-2~\citep{socher2013recursive}, MR~\citep{pang2005seeing}, and CR~\citep{hu2004mining}. The task is to classify each sentence as \emph{positive} or \emph{negative}.
    \item \textbf{Multiclass sentiment classification:} SST-5~\citep{socher2013recursive}, which extends sentiment analysis to five labels: \emph{terrible}, \emph{bad}, \emph{okay}, \emph{good}, and \emph{great}.
    \item \textbf{Question classification:} TREC~\citep{voorhees2000building}, where each question is categorized into one of six classes: \emph{Description}, \emph{Entity}, \emph{Expression}, \emph{Human}, \emph{Location}, or \emph{Number}.
    \item \textbf{News classification:} AG's News~\citep{zhang2015character}, which assigns each news article to one of four topics: \emph{World}, \emph{Sports}, \emph{Business}, or \emph{Tech}.
    \item \textbf{Subjectivity classification:} SUBJ~\citep{pang2004sentimental}, where the goal is to determine whether a sentence is \emph{subjective} or \emph{objective}.
\end{itemize}

We run each method three times and report mean accuracy with standard deviations.
Results are shown in Table~\ref{tab:cls}.
Overall, \ours{} achieves the best average performance across the seven classification benchmarks (84.19), despite using no task-specific training data and no dataset-level prompt optimization.

Across individual datasets, \ours{} attains the top result on three benchmarks: TREC (86.15), SUBJ (82.18), and MR (91.78), while remaining competitive on the remaining tasks (e.g., 96.18 on SST-2 and 85.61 on AG's News). The largest improvement is on TREC, where \ours{} outperforms the strongest competing method by roughly seven points. Gains on SUBJ and MR are smaller but consistent, supporting the conclusion that instance-specific example synthesis combined with paraphrase-and-vote aggregation can improve robustness without supervision. Example prompts are provided in Appendix~\ref{sec:appendix_prompts}.

\paragraph{Mathematical Reasoning.}
We evaluate mathematical reasoning on three benchmarks: GSM8K~\citep{cobbe2021gsm8k}, a dataset of grade-school word problems requiring multi-step arithmetic; DeepMath~\citep{deepmath}, a decontaminated benchmark of harder, verifiable problems; and MATH500~\citep{math500}, a 500-problem subset of the MATH benchmark covering competition-style questions.

Table~\ref{tab:reasoning} shows that \ours{} achieves the best accuracy among all compared methods on GSM8K (94.67\%) and DeepMath (27.43\%), improving over the strongest baseline by +2.55 and +2.10 points, respectively. Importantly, these gains are obtained without any benchmark-specific prompt optimization, whereas several baselines explicitly search for or tune prompts on the target dataset (and GPS is optimized on GSM8K/DeepMath in its standard setup). On MATH500, where GPS and \ours{} operate out-of-distribution, \ours{} remains competitive (42.47\%), outperforming instruction-optimization methods such as APE, GA, and GRACE as well as both GPS variants, though it trails PRL and PIAST. Overall, these results suggest that per-instance prompt construction, combining synthesized in-context examples with paraphrase-and-vote aggregation, can substantially improve reasoning performance without expensive dataset-level optimization loops.

\paragraph{Domain knowledge task.}
To assess performance on tasks that require domain-specific expertise, we additionally evaluate \ours{} on MedQA~\citep{medqa}, a multiple-choice medical question answering benchmark with four options (A-D). Results are reported in Table~\ref{tab:reasoning}. Notably, GPS has been trained on MedQA, whereas \ours{} does not use any MedQA-specific data.
In this sense, \ours{} is the only out-of-distribution method in the table for MedQA. Despite this constraint, \ours{} remains competitive with some instruction-optimization baselines (e.g., APE), highlighting that per-instance example synthesis can transfer effectively to specialized domains.

\begin{table}[ht]
\centering
\renewcommand{\arraystretch}{1.0}
\setlength{\tabcolsep}{4pt}
\caption{Results on reasoning benchmarks (higher is better). Numbers are mean accuracy with standard deviation.
Colours mark the best (\redc{red}), second-best (\orangec{orange}) and third-best (\yellowc{yellow}) methods in each column.}
\resizebox{1.0\columnwidth}{!}{%
\begin{tabular}{lcccc}
\hline
\textbf{Method} & \textbf{GSM8K} & \textbf{DeepMath} & \textbf{MATH500} & \textbf{MedQA} \\
\hline
APE
& 83.43\textsubscript{\textcolor{gray}{$\pm$1.98}}
& 15.47\textsubscript{\textcolor{gray}{$\pm$0.45}}
& 31.53\textsubscript{\textcolor{gray}{$\pm$1.04}}
& 45.66\textsubscript{\textcolor{gray}{$\pm$0.97}} \\
GA
& 81.62\textsubscript{\textcolor{gray}{$\pm$1.38}}
& 18.63\textsubscript{\textcolor{gray}{$\pm$2.37}}
& 40.13\textsubscript{\textcolor{gray}{$\pm$1.39}}
& 51.95\textsubscript{\textcolor{gray}{$\pm$1.61}} \\
DE
& 79.52\textsubscript{\textcolor{gray}{$\pm$0.45}}
& 16.10\textsubscript{\textcolor{gray}{$\pm$0.00}}
& 34.20\textsubscript{\textcolor{gray}{$\pm$1.39}}
& 51.76\textsubscript{\textcolor{gray}{$\pm$0.16}} \\
GRACE
& 82.37\textsubscript{\textcolor{gray}{$\pm$1.82}}
& 15.05\textsubscript{\textcolor{gray}{$\pm$0.16}}
& 33.20\textsubscript{\textcolor{gray}{$\pm$1.60}}
& 52.26\textsubscript{\textcolor{gray}{$\pm$0.16}} \\
PRL
& 86.15\textsubscript{\textcolor{gray}{$\pm$0.55}}
& 21.58\textsubscript{\textcolor{gray}{$\pm$0.22}}
& \yellowc{44.40\textsubscript{\textcolor{gray}{$\pm$1.40}}}
& \orangec{53.34\textsubscript{\textcolor{gray}{$\pm$0.11}}} \\
PIAST
& \yellowc{91.65\textsubscript{\textcolor{gray}{$\pm$0.31}}}
& \yellowc{24.85\textsubscript{\textcolor{gray}{$\pm$1.13}}}
& \orangec{48.53\textsubscript{\textcolor{gray}{$\pm$0.31}}}
& 52.45\textsubscript{\textcolor{gray}{$\pm$0.51}} \\
PIAST (E)
& \orangec{92.12\textsubscript{\textcolor{gray}{$\pm$0.12}}}
& \orangec{25.33\textsubscript{\textcolor{gray}{$\pm$0.75}}}
& \redc{48.70\textsubscript{\textcolor{gray}{$\pm$0.50}}}
& 52.89\textsubscript{\textcolor{gray}{$\pm$0.05}} \\
GPS-SR-0.1
& 87.55\textsubscript{\textcolor{gray}{$\pm$0.42}}
& 21.58\textsubscript{\textcolor{gray}{$\pm$0.22}}
& 31.80\textsubscript{\textcolor{gray}{$\pm$1.60}}
& \yellowc{53.31\textsubscript{\textcolor{gray}{$\pm$1.47}}} \\
GPS-J
& 84.45\textsubscript{\textcolor{gray}{$\pm$0.93}}
& 21.40\textsubscript{\textcolor{gray}{$\pm$0.25}}
& 34.20\textsubscript{\textcolor{gray}{$\pm$0.80}}
& \redc{54.92\textsubscript{\textcolor{gray}{$\pm$0.14}}} \\
\ours{}
& \redc{\textbf{94.95}\textsubscript{\textcolor{gray}{$\pm$0.30}}}
& \redc{\textbf{27.78}\textsubscript{\textcolor{gray}{$\pm$0.40}}}
& 43.20\textsubscript{\textcolor{gray}{$\pm$0.33}}
& 45.64\textsubscript{\textcolor{gray}{$\pm$1.34}} \\
\hline
\end{tabular}%
}
\label{tab:reasoning}
\end{table}

\subsection{Ablation Studies}
\label{sec:ablations}

In this section, we perform ablation experiments to study the robustness and scalability of TATRA. We focus on three questions: (1) Does TATRA generalize across different model families? (2) How does performance scale with the size of the evaluator model? (3) What is the sensitivity of the method to hyperparameter configurations?

\paragraph{Cross-Model Generalization.}
TATRA decouples the example generator $\mathcal{M}_{\text{GEN}}$ from the prompt evaluator $\mathcal{M}_{\text{EVAL}}$. 
To verify that our method does not rely on capabilities shared only between identical models, we evaluate performance when mixing model families, specifically \texttt{Llama-3.1-8B-Instruct} and \texttt{Qwen2.5-Instruct-7B}.

As shown in Table~\ref{tab:cross_model}, cross-model pairings yield robust performance. Notably, the \texttt{Llama8B} $\rightarrow$ \texttt{Qwen7B} configuration (using \texttt{Llama} for generation and \texttt{Qwen} for evaluation) achieves 95.5\% on GSM8K and 43.0\% on MATH500. This significantly outperforms the homogeneous \texttt{Llama} $\rightarrow$ \texttt{Llama} baseline (83.3\% on GSM8K), suggesting that the synthetic examples generated by TATRA are universally effective and transferrable, rather than being artifacts tailored to a specific model's distribution. This confirms that a stronger evaluator can effectively leverage examples produced by a weaker generator.

\begin{table}[t]
    \centering
    \caption{Cross-model pairing study. Notation denotes Generator $\rightarrow$ Evaluator. Results are averaged over three seeds.}
    \label{tab:cross_model}
    \resizebox{\columnwidth}{!}{%
    \begin{tabular}{lcccc}
        \toprule
        Configuration & GSM8K & DeepMath & MATH500 & MedQA \\
        \midrule
        \texttt{Qwen} $\rightarrow$ \texttt{Qwen} & 95.0 & \textbf{27.8} & \textbf{43.2} & \textbf{45.64}  \\
        \texttt{Llama} $\rightarrow$ \texttt{Llama} & 83.3 & 26.3 & 39.2 & 21.3 \\
        \texttt{Llama} $\rightarrow$ \texttt{Qwen}  & \textbf{95.5} & \textbf{27.8} & 43.0 & 28.1 \\
        \texttt{Qwen} $\rightarrow$ \texttt{Llama}  & 87.5 & 26.4 & 40.5 & 33.6 \\
        \bottomrule
    \end{tabular}%
    }

\end{table}

\begin{table}[h]
    \centering
    \caption{Benchmark accuracy (\%) across evaluator sizes. The generator is fixed as \texttt{Qwen2.5-Instruct-7B}.}
    \label{tab:evaluator_scaling}
    \begin{small}
    \begin{sc}
    \resizebox{\columnwidth}{!}{%
    \begin{tabular}{lcccc}
        \toprule
        Evaluator Size & GSM8K & DeepMath & MATH500 & MedQA \\
        \midrule
        3B  & 86.5 & 26.4 & 34.2 & 36.1 \\
        7B (Ours) & 94.7 & 27.4 & 42.5 & 44.6 \\
        14B & \textbf{96.2} & \textbf{28.0} & \textbf{40.0} & \textbf{50.7} \\
        \bottomrule
    \end{tabular}%
    }
    \end{sc}
    \end{small}
\end{table}

\paragraph{Scaling Evaluator Size.}
We investigate how TATRA scales with the capacity of the evaluator model. Keeping the generator fixed at \texttt{Qwen2.5-Instruct-7B}, we vary the evaluator size across 3B, 7B, and 14B parameters.

Table~\ref{tab:evaluator_scaling} illustrates a clear scaling trend: larger evaluators consistently translate to higher downstream accuracy. Even with a compact 3B evaluator, TATRA maintains strong performance, achieving 86.5\% on GSM8K, demonstrating the method's viability for resource-constrained inference. Scaling to 14B yields substantial gains, boosting MedQA accuracy from 36.1\% (3B) to 50.7\% and GSM8K to 96.2\%. 
These results indicate effectiveness of \ours{} across LLM sizes.

\paragraph{Hyperparameter Sensitivity.}
We study the sensitivity of \ours{} to two key hyperparameters: the number of paraphrases $n$ and the number of synthesized in-context examples $k$.
We perform a grid search on DeepMath and MATH500, varying $n \in \{0,\dots,15\}$ and $k \in \{4,\dots,32\}$.

Overall, performance exhibits limited sensitivity to $k$, with consistently strong results for intermediate values. In particular, we observe a favorable operating regime for $k \in \{8,16\}$. For example, on MATH500 with $n=5$, accuracy reaches its maximum at $k=16$ (43.3\%). Increasing the number of paraphrases generally improves robustness by reducing variance due to prompt surface form, but the gains diminish beyond $n=5$.
Indeed, further increasing $n$ up to 15 yields minimal additional improvement (e.g., DeepMath accuracy remains approximately 27.8\%).
Based on these results, we set $n=5$ and $k=8$ as default values, as they provide a favorable trade-off between computational cost and performance.

\begin{figure}[t]
    \centering
    \begin{subfigure}{0.49\linewidth}
        \includegraphics[width=\linewidth]{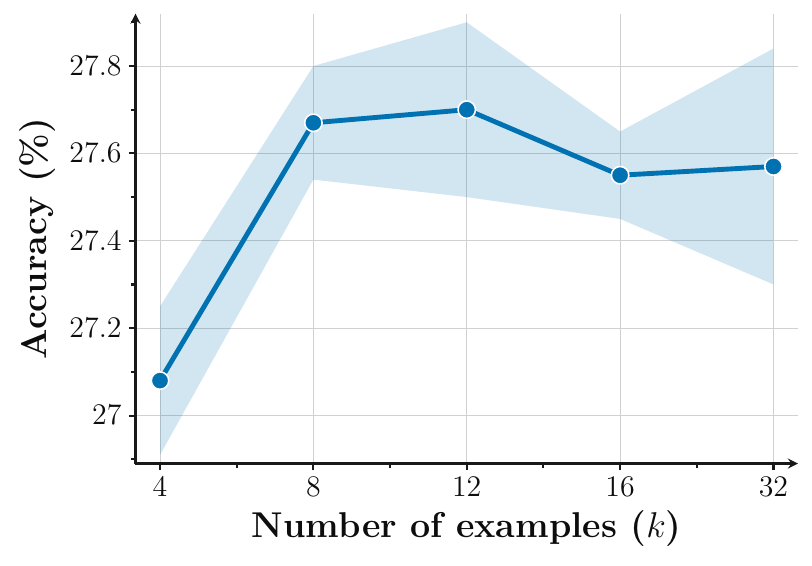}
        \caption{DeepMath}
    \end{subfigure}
    \hfill
    \begin{subfigure}{0.49\linewidth}
        \includegraphics[width=\linewidth]{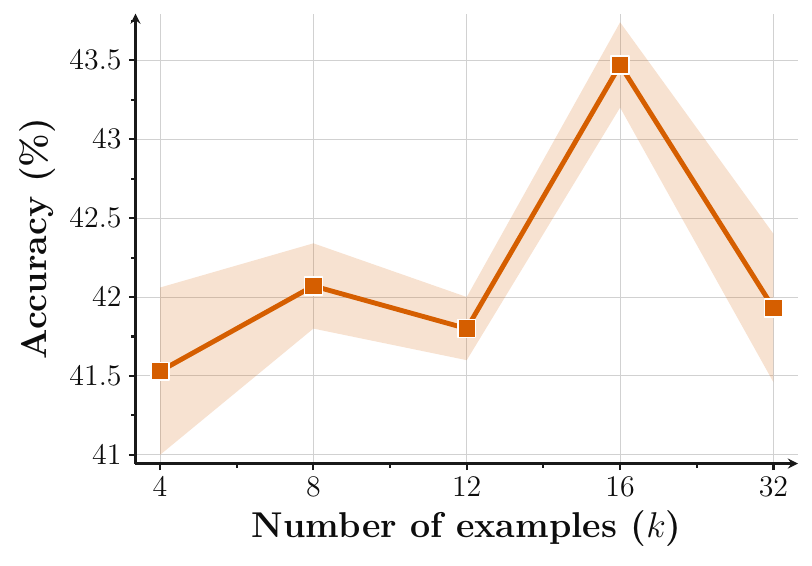}
        \caption{MATH500}
    \end{subfigure}

    \caption{\textbf{Impact of increasing in-context examples ($k$).} We vary $k \in \{4, \dots, 32\}$ while keeping $n$ fixed. The shading indicates the recorded seed std. deviation}
    \label{fig:k_ablation}
\end{figure}

\begin{figure}[t]
    \centering
    \begin{subfigure}{0.49\linewidth}
        \includegraphics[width=\linewidth]{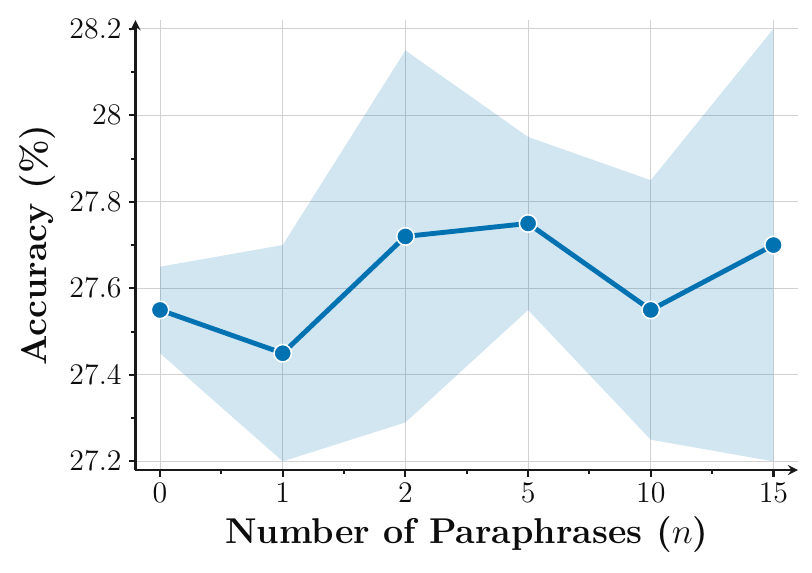}
        \caption{DeepMath}
    \end{subfigure}
    \hfill
    \begin{subfigure}{0.49\linewidth}
        \includegraphics[width=\linewidth]{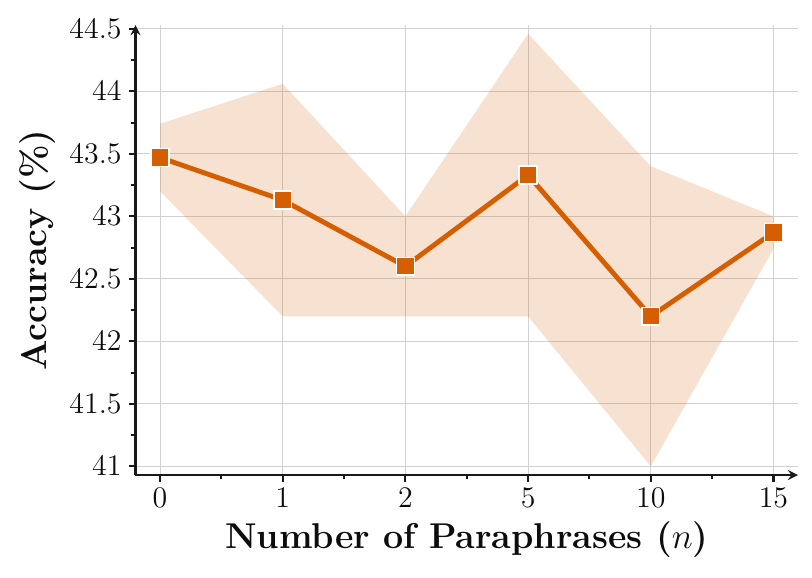}
        \caption{MATH500}
    \end{subfigure}

    \caption{\textbf{Impact of increasing paraphrases ($n$).} We vary $n \in \{0, \dots, 15\}$ while keeping $k$ fixed. The shading indicates the recorded seed std. deviation.}
    \label{fig:n_ablation}
\end{figure}

\paragraph{Compute and runtime.}
We report wall-clock runtimes of \ours{} in Appendix~\ref{app:runtime} (Table~\ref{tab:runtime}). Runtime differences are dominated by dataset size and average sequence length (and, for reasoning tasks, longer generations), rather than by any task-specific optimization. Most classification benchmarks finish within $\sim$10-45 minutes, whereas NEWS, DeepMath, and GSM8K incur higher cost due to longer inputs and/or more tokens generated. The small seed-to-seed deviations suggest that runtime is stable under different random seeds.

\section{Conclusions \& Limitations}
We have shown that automatic prompt engineering without access to any ground truth can outperform methods that get training signals from labeled datasets.
This is surprising, since typically in machine learning access to a training set with ground truth labels usually improves performance.
We argue that our setting is also more realistic, since only rarely do practitioners have access to labeled datasets for their task at hand.
We hope to inspire future work on automatic prompting without training set access. A limitation of our method is its increased compute for each sample: We perform instance-adaptive prompt construction and aggregation, which requires rerunning the generation, paraphrasing, and voting pipeline separately for each new observation. In contrast, most automated prompt engineering approaches optimize a single dataset-level prompt once and then reuse it for fast inference on new examples without per-instance regeneration.
On the other hand we do not incur any training time overhead that methods with training-set access may have.

%

\nocite{langley00}

\bibliographystyle{icml2026}

\newpage
\appendix
\onecolumn
\section{Hyperparameter choice} \label{sec:appendix_hyperparams}
\begin{promptbox}{Hyperparameters (fixed test runs)}
Fixed hyperparameters used for all standard test runs (excluding hyperparameter exploration):

- paraphrases per item: n = 10
- in-context examples:  k = 16
- repeated runs:        r = 15
\end{promptbox}

We select the fixed hyperparameter setting \((n,k,r)=(10,16,15)\) for standard test runs based on the hyperparameter sensitivity analysis: among the explored configurations, this choice yields the highest average performance when results are aggregated across all runs and all random seeds. Concretely, we prioritize configurations that maximize mean accuracy under repeated paraphrase sampling and repeated runs, because higher averages indicate that the proposed procedure is being applied effectively and consistently rather than relying on chance.

This selection is also aligned with the intent of the method: the goal is not to minimize paraphrases, demonstrations, or repeats, but to use them sufficiently to stabilize predictions via majority voting and to reduce variance across seeds. Therefore, we adopt the configuration that performs best on average under the same evaluation protocol used throughout the paper.

\section{Prompt templates} \label{sec:appendix_prompts}

This appendix reports the exact prompt templates used to synthesize in-context examples and paraphrases for each benchmark in our experiments. We include them to ensure full reproducibility and to clarify the task-specific constraints imposed on the generator model (e.g., strict output format, label-specific style rules, and topic/sentence-length diversity plans).

All templates contain placeholders (e.g., \texttt{\{k\}}, \texttt{\{label\}}, \texttt{\{topics\}}, \texttt{\{counts\}}) that are instantiated at runtime from the task configuration and the chosen hyperparameters. Unless stated otherwise, we require the generator to output only examples in the specified format (no additional commentary) so that outputs can be parsed deterministically.

\subsection{Example generation prompts for tasks}

\begin{promptbox}{CR (example generation)}
You are a data generator that writes high-quality in-context learning examples for classification.

Create exactly {num_examples} training examples in THIS STRICT format only:

Example1:
Sentence: "<text>"
Label: {label}
...
Example{num_examples}:
Sentence: "<text>"
Label: {label}

Diversity plan (MUST FOLLOW):
- Example1: write exactly {counts[0]} sentence(s); topic: {topics[0]}.
- Example2: write exactly {counts[1]} sentence(s); topic: {topics[1]}.
- ...
(Topics are chosen from TASK_CONFIGS["cr"]["topics"][{label}].)

Rules:
- Each example's "Sentence" must contain exactly the number of sentences specified above (1-3).
- Keep sentences concise: typically 3-14 words each.
- Use only ASCII characters. Do NOT include double quotes inside the text.
- Do NOT wrap output in Markdown/code fences.
- Apply label-specific style rules from TASK_CONFIGS["cr"]["style_rules"][{label}].
- Output ONLY the examples in the exact format above; no extra text.
\end{promptbox}

\begin{promptbox}{MR (example generation)}
You are a data generator that writes high-quality in-context learning examples for classification.

Create exactly {num_examples} training examples in THIS STRICT format only:

Example1:
Sentence: "<text>"
Label: {label}
...
Example{num_examples}:
Sentence: "<text>"
Label: {label}

Diversity plan (MUST FOLLOW):
- Example1: write exactly {counts[0]} sentence(s); topic: {topics[0]}.
- Example2: write exactly {counts[1]} sentence(s); topic: {topics[1]}.
- ...
(Topics are chosen from TASK_CONFIGS["mr"]["topics"][{label}].)

Rules:
- Each example's "Sentence" must contain exactly the number of sentences specified above (1-3).
- Keep sentences concise: typically 3-14 words each.
- Use only ASCII characters. Do NOT include double quotes inside the text.
- Do NOT wrap output in Markdown/code fences.
- Apply label-specific style rules from TASK_CONFIGS["mr"]["style_rules"][{label}].
- Output ONLY the examples in the exact format above; no extra text.
\end{promptbox}

\begin{promptbox}{SST2 (example generation)}
You are a data generator that writes high-quality in-context learning examples for classification.

Create exactly {num_examples} training examples in THIS STRICT format only:

Example1:
Sentence: "<text>"
Label: {label}
...
Example{num_examples}:
Sentence: "<text>"
Label: {label}

Diversity plan (MUST FOLLOW):
- Example1: write exactly {counts[0]} sentence(s); topic: {topics[0]}.
- Example2: write exactly {counts[1]} sentence(s); topic: {topics[1]}.
- ...
(Topics come from a sentiment topic pool; the correct label is fixed per run.)

Rules:
- Each example's "Sentence" must contain exactly the number of sentences specified above (1-3).
- Keep sentences concise: typically 3-14 words each.
- Use only ASCII characters. Do NOT include double quotes inside the text.
- Use exactly ONE 'Sentence:' line per example; if multiple sentences are needed, put them inside the same quotes separated by a space.
- Do NOT wrap output in Markdown/code fences.
- Label MUST be exactly: {label} (one of: positive/negative).
- If {label}=positive: write clearly POSITIVE sentiment (praise, enjoyment, recommendation).
- If {label}=negative: write clearly NEGATIVE sentiment (criticism, disappointment, complaints).
- Output ONLY the examples in the exact format above; no extra text.
\end{promptbox}

\begin{promptbox}{SST5 (example generation)}
You are a data generator that writes high-quality in-context learning examples for classification.

Create exactly {num_examples} training examples in THIS STRICT format only:

Example1:
Sentence: "<text>"
Label: {label}
...
Example{num_examples}:
Sentence: "<text>"
Label: {label}

Diversity plan (MUST FOLLOW):
- Example1: write exactly {counts[0]} sentence(s); topic: acting.
- Example2: write exactly {counts[1]} sentence(s); topic: plot.
- ...
(Topics cycle through: acting; plot; direction; dialogue; pacing.)

Rules:
- Each example's "Sentence" must contain exactly the number of sentences specified above (1-3).
- Keep sentences concise: typically 3-14 words each.
- Use only ASCII characters. Do NOT include double quotes inside the text.
- Do NOT wrap output in Markdown/code fences.
- Apply label-intensity rules (terrible/bad/okay/good/great) as specified in TASK_CONFIGS["sst5"]["style_rules"][{label}].
- Output ONLY the examples in the exact format above; no extra text.
\end{promptbox}

\begin{promptbox}{NEWS (example generation)}
You are a data generator that writes high-quality in-context learning examples for classification.

Create exactly {num_examples} training examples in THIS STRICT format only:

Example1:
Sentence: "<text>"
Label: {label}
...
Example{num_examples}:
Sentence: "<text>"
Label: {label}

Diversity plan (MUST FOLLOW):
- Example1: write exactly {counts[0]} sentence(s); topic: {topics[0]}.
- Example2: write exactly {counts[1]} sentence(s); topic: {topics[1]}.
- ...
(Topics are chosen from TASK_CONFIGS["news"]["topics"][{label}].)

Rules:
- Each example's "Sentence" must contain exactly the number of sentences specified above (1-3).
- Keep sentences concise: typically 3-14 words each.
- Use only ASCII characters. Do NOT include double quotes inside the text.
- Do NOT wrap output in Markdown/code fences.
- Apply label-specific style rules from TASK_CONFIGS["news"]["style_rules"][{label}].
- Output ONLY the examples in the exact format above; no extra text.
\end{promptbox}

\begin{promptbox}{TREC (example generation)}
You are a data generator that writes high-quality in-context learning examples for classification.

Create exactly {num_examples} training examples in THIS STRICT format only:

Example1:
Sentence: "<text>"
Label: {label}
...
Example{num_examples}:
Sentence: "<text>"
Label: {label}

Diversity plan (MUST FOLLOW):
- Example1: write exactly {counts[0]} sentence(s); topic: {topics[0]}.
- Example2: write exactly {counts[1]} sentence(s); topic: {topics[1]}.
- ...
(Topics are chosen from TASK_CONFIGS["trec"]["topics"][{label}].)

Rules:
- Each example's "Sentence" must contain exactly the number of sentences specified above (1-3).
- Keep sentences concise: typically 3-14 words each.
- Use only ASCII characters. Do NOT include double quotes inside the text.
- Do NOT wrap output in Markdown/code fences.
- Apply label-specific style rules from TASK_CONFIGS["trec"]["style_rules"][{label}].
- Output ONLY the examples in the exact format above; no extra text.
\end{promptbox}

\begin{promptbox}{SUBJ (example generation)}
You generate training examples for subjectivity classification. SUBJECTIVE = opinions about the movie. OBJECTIVE = factual plot/story descriptions.

Create exactly {example_count} training examples in this exact format:

Example1:
Sentence: "<text>"
Label: {target_label}

Example2:
Sentence: "<text>"
Label: {target_label}

...

Diversity plan (follow exactly):
- Example1: {sentence_counts[0]} sentence(s); topic: {topic_list[0]}.
- Example2: {sentence_counts[1]} sentence(s); topic: {topic_list[1]}.
- ...

Rules:
- Each Sentence must contain exactly the specified number of sentences (1-3).
- Keep sentences concise: 3-14 words each.
- Include variety: at least one short (<=5 words) and one longer (10-14 words).
- Use only ASCII characters, no quotes inside the text.
- Put multiple sentences in the same quotes separated by spaces.
- No markdown or code fences.
- If {target_label}=subjective: write opinions/judgments (evaluative language).
- If {target_label}=objective: write factual plot/story descriptions (no judgment).
- Output ONLY the examples, no extra text.
\end{promptbox}

\begin{promptbox}{MedQA (example generation)}
System prompt
You are a data generator that writes high-quality medical multiple-choice questions for in-context learning.

Example generation template
Create exactly {k} training examples in THIS STRICT format only:

Example1:
Question: "<medical question with 5 options A-E>"
Answer: {label}

Example2:
Question: "<medical question with 5 options A-E>"
Answer: {label}

...
Example{k}:
Question: "<medical question with 5 options A-E>"
Answer: {label}

Diversity plan (MUST FOLLOW):
{plan}

Rules:
- Each question MUST include options A:, B:, C:, D:, E:
- The correct answer is always: {label}
- Use realistic medical scenarios (diagnosis, treatment, pharmacology)
- Use only ASCII characters
- Do NOT wrap output in Markdown/code fences
- Output ONLY the examples in the exact format above; no extra text

Example categories (topics)
For each label A-E: diagnosis; treatment; pharmacology; pathology; anatomy.
\end{promptbox}

\begin{promptbox}{GSM8K (example generation)}
System prompt
You are a data generator that writes grade-school math word problems with integer answers.

Example generation template
Create exactly {k} training examples in THIS STRICT format only:

Example1:
Question: "<math problem>"
Answer: <answer>

Example2:
Question: "<math problem>"
Answer: <answer>

...
Example{k}:
Question: "<math problem>"
Answer: <answer>

Diversity plan (MUST FOLLOW):
{plan}

Rules:
- Each question must be a well-formed math problem
- Each answer must be the correct solution
- Problems should be simple word problems with clear numerical answers
- Answers MUST be integers
- Use only ASCII characters (except for TeX math symbols)
- Do NOT wrap output in Markdown/code fences
- Output ONLY the examples in the exact format above; no extra text

Example categories (topics)
arithmetic; algebra; word problems; percentages; ratios.
\end{promptbox}

\begin{promptbox}{DeepMath (example generation)}
System prompt
You are a data generator that writes advanced mathematics problems.

Example generation template
Create exactly {k} training examples in THIS STRICT format only:

Example1:
Question: "<math problem>"
Answer: <answer>

Example2:
Question: "<math problem>"
Answer: <answer>

...
Example{k}:
Question: "<math problem>"
Answer: <answer>

Diversity plan (MUST FOLLOW):
{plan}

Rules:
- Each question must be a well-formed math problem
- Each answer must be the correct solution
- Problems can involve calculus, linear algebra, number theory, probability
- Use TeX notation for math expressions
- Use only ASCII characters (except for TeX math symbols)
- Do NOT wrap output in Markdown/code fences
- Output ONLY the examples in the exact format above; no extra text

Example categories (topics)
number theory; calculus; linear algebra; probability; combinatorics.
\end{promptbox}

\begin{promptbox}{Math500 (example generation)}
System prompt
You are a data generator that writes competition-style mathematics problems.

Example generation template
Create exactly {k} training examples in THIS STRICT format only:

Example1:
Question: "<math problem>"
Answer: <answer>

Example2:
Question: "<math problem>"
Answer: <answer>

...
Example{k}:
Question: "<math problem>"
Answer: <answer>

Diversity plan (MUST FOLLOW):
{plan}

Rules:
- Each question must be a well-formed math problem
- Each answer must be the correct solution
- Problems should be challenging but solvable
- Use TeX notation for math expressions
- Use only ASCII characters (except for TeX math symbols)
- Do NOT wrap output in Markdown/code fences
- Output ONLY the examples in the exact format above; no extra text

\textbf{Example categories (topics)}\par
algebra; geometry; number theory; counting; precalculus.
\end{promptbox}

\subsection{Paraphrase generation prompts for tasks}

\begin{promptbox}{CR (paraphrase generation)}
You paraphrase product review sentences while preserving the original sentiment and meaning.
Produce exactly {n} diverse paraphrases of the sentence below. Keep tone and sentiment identical.
Output ONLY the paraphrases, one per line, with no numbering, bullets, or extra commentary.

Sentence:
{s}
\end{promptbox}

\begin{promptbox}{MR (paraphrase generation)}
You paraphrase movie-review sentences while preserving the original sentiment and meaning.
Produce exactly {n} diverse paraphrases of the sentence below. Keep tone and sentiment identical.
Output ONLY the paraphrases, one per line, with no numbering, bullets, or extra commentary.

Sentence:
{s}
\end{promptbox}

\begin{promptbox}{SST2 (paraphrase generation)}
You paraphrase movie-review sentences while preserving the original sentiment and meaning.
Produce exactly {n} diverse paraphrases of the sentence below. Keep tone and sentiment identical.
Output ONLY the paraphrases, one per line, with no numbering, bullets, or extra commentary.

Sentence:
{s}
\end{promptbox}

\begin{promptbox}{SST5 (paraphrase generation)}
You paraphrase movie-review sentences while preserving the original sentiment intensity.
Produce exactly {n} diverse paraphrases. Keep the sentiment level (terrible/bad/okay/good/great) identical.
Output ONLY the paraphrases, one per line.

Sentence:
{s}
\end{promptbox}

\begin{promptbox}{NEWS (paraphrase generation)}
You paraphrase news headlines while preserving the original topic and meaning.
Produce exactly {n} diverse paraphrases. Keep the topic category identical.
Output ONLY the paraphrases, one per line.

Headline:
{s}
\end{promptbox}

\begin{promptbox}{TREC (paraphrase generation)}
You paraphrase questions while preserving the question type and intent.
Produce exactly {n} diverse paraphrases. Keep the answer type requirement identical.
Output ONLY the paraphrases, one per line.

Question:
{s}
\end{promptbox}

\begin{promptbox}{SUBJ (paraphrase generation)}
You paraphrase sentences while preserving the original MEANING and SUBJECTIVITY TYPE (whether it expresses opinion or describes facts).
Produce exactly {n} diverse paraphrases of the sentence below. Keep the subjectivity classification identical.
Output ONLY the paraphrases, one per line, without numbering or commentary.

Sentence:
{s}
\end{promptbox}

\begin{promptbox}{MedQA (paraphrase generation)}
You rephrase medical questions while preserving the clinical scenario and answer options.
Produce exactly {n} diverse rephrasings of the question below. Keep the medical meaning identical.
Output ONLY the rephrasings, one per line, with no numbering, bullets, or extra commentary.

Question:
{s}
\end{promptbox}

\begin{promptbox}{GSM8K (paraphrase generation)}
You rephrase math word problems while preserving the mathematical structure and answer.
Produce exactly {n} diverse rephrasings of the problem below. Keep numbers and relationships identical.
Output ONLY the rephrasings, one per line, with no numbering, bullets, or extra commentary.

Problem:
{s}
\end{promptbox}

\begin{promptbox}{DeepMath (paraphrase generation)}
You rephrase advanced mathematics problems while preserving the mathematical content.
Produce exactly {n} diverse rephrasings of the problem below. Keep notation and meaning identical.
Output ONLY the rephrasings, one per line, with no numbering, bullets, or extra commentary.

Problem:
{s}
\end{promptbox}

\begin{promptbox}{Math500 (paraphrase generation)}
You rephrase competition mathematics problems while preserving the mathematical content.
Produce exactly {n} diverse rephrasings of the problem below. Keep notation and meaning identical.
Output ONLY the rephrasings, one per line, with no numbering, bullets, or extra commentary.

Problem:
{s}
\end{promptbox}

\subsection{Topic pools for tasks}\label{app:topicpools}

\begin{promptbox}{SUBJ (topic pool)}
SUBJECTIVE_TOPICS =
- opinion on acting
- judgment of direction
- critique of dialogue
- assessment of cinematography
- evaluation of pacing
- review of soundtrack
- opinion on visual effects
- judgment of set design
- assessment of tone
- opinion on themes
- evaluation of casting
- critique of character development
- assessment of humor
- emotional reaction

OBJECTIVE_TOPICS =
- plot summary
- character actions
- story events
- character relationships
- plot twist description
- setting description
- character background
- story conflict
- character motivation
- narrative arc
- scene description
- character introduction
- plot setup
- story resolution
\end{promptbox}

\begin{promptbox}{SST2 (topic pool)}
SENTIMENT_TOPICS =
- acting/performance
- direction
- screenplay/dialogue
- cinematography
- editing
- pacing
- soundtrack/music
- visual effects
- set & costume design
- genre/tone
- themes/message
- casting choices
- character development
- humor
- emotional impact
\end{promptbox}

\begin{promptbox}{TREC (topic pool)}
topics["Abbreviation"] =
- acronym meaning
- abbreviation expansion

topics["Entity"] =
- product
- animal
- color
- invention
- food

topics["Description"] =
- definition
- manner
- reason

topics["Human"] =
- person name
- inventor
- author
- discoverer

topics["Location"] =
- city
- country
- mountain
- address

topics["Number"] =
- date
- count
- distance
- money
- percentage
\end{promptbox}

\begin{promptbox}{NEWS (topic pool)}
topics["World"] =
- international politics
- war/conflict
- diplomacy
- elections
- human rights

topics["Sports"] =
- football
- basketball
- olympics
- tennis
- soccer

topics["Business"] =
- stock market
- mergers
- economy
- earnings
- banking

topics["Tech"] =
- software
- hardware
- internet
- AI
- startups
\end{promptbox}

\begin{promptbox}{MR (topic pool)}
topics["positive"] =
- acting/performance
- direction
- screenplay/dialogue
- cinematography
- pacing
- soundtrack/music
- visual effects
- themes/message
- character development
- emotional impact

topics["negative"] =
- acting/performance
- direction
- screenplay/dialogue
- cinematography
- pacing
- soundtrack/music
- visual effects
- themes/message
- character development
- emotional impact
\end{promptbox}

\begin{promptbox}{CR (topic pool)}
topics["positive"] =
- product quality
- customer service
- value for money
- durability
- ease of use
- design
- performance
- reliability
- features
- shipping

topics["negative"] =
- product defects
- poor service
- overpriced
- fragility
- complexity
- ugly design
- slow performance
- unreliable
- missing features
- shipping issues
\end{promptbox}

\begin{promptbox}{SST5 (topic pool)}
topics["terrible"] =
- acting
- plot
- direction
- dialogue
- pacing

topics["bad"] =
- acting
- plot
- direction
- dialogue
- pacing

topics["okay"] =
- acting
- plot
- direction
- dialogue
- pacing

topics["good"] =
- acting
- plot
- direction
- dialogue
- pacing

topics["great"] =
- acting
- plot
- direction
- dialogue
- pacing
\end{promptbox}

\begin{promptbox}{MedQA (topic pool)}
topics["A"] =
- diagnosis
- treatment
- pharmacology
- pathology
- anatomy

topics["B"] =
- diagnosis
- treatment
- pharmacology
- pathology
- anatomy

topics["C"] =
- diagnosis
- treatment
- pharmacology
- pathology
- anatomy

topics["D"] =
- diagnosis
- treatment
- pharmacology
- pathology
- anatomy

topics["E"] =
- diagnosis
- treatment
- pharmacology
- pathology
- anatomy
\end{promptbox}

\begin{promptbox}{DeepMath (topic pool)}
topics =
- number theory
- calculus
- linear algebra
- probability
- combinatorics
\end{promptbox}

\begin{promptbox}{GSM8K (topic pool)}
topics =
- arithmetic
- algebra
- word problems
- percentages
- ratios
\end{promptbox}

\begin{promptbox}{Math500 (topic pool)}
topics =
- algebra
- geometry
- number theory
- counting
- precalculus
\end{promptbox}

\section{Hyperparameter sensitivity analysis results}

\begingroup
\footnotesize
\setlength{\tabcolsep}{3.6pt}
\renewcommand{\arraystretch}{0.85}

\begin{longtable}{lcc@{\hspace{10pt}}cc}
\caption{\textbf{Hyperparameter Sensitivity Analysis ($n, k$).} Comparison of DeepMath and MATH500 performance across varying configuration parameters. We report the Mean and Deviation across 3 seeds.}
\label{tab:hyperparams}\\

\toprule
 & \multicolumn{2}{c}{\textbf{DEEPMATH}} & \multicolumn{2}{c}{\textbf{MATH500}} \\
\cmidrule(lr){2-3}\cmidrule(lr){4-5}
\textbf{Config ($n,k$)} & \textbf{Mean} & \textbf{Dev.} & \textbf{Mean} & \textbf{Dev.} \\
\midrule
\endfirsthead

\toprule
 & \multicolumn{2}{c}{\textbf{DEEPMATH}} & \multicolumn{2}{c}{\textbf{MATH500}} \\
\cmidrule(lr){2-3}\cmidrule(lr){4-5}
\textbf{Config ($n,k$)} & \textbf{Mean} & \textbf{Dev.} & \textbf{Mean} & \textbf{Dev.} \\
\midrule
\endhead

\midrule
\multicolumn{5}{r}{\emph{Continued on next page}}\\
\endfoot

\bottomrule
\endlastfoot

$n=0, k=4$  & 27.08 & 0.17 & 41.53 & 0.53 \\
$n=0, k=8$  & 27.67 & 0.13 & 42.07 & 0.27 \\
$n=0, k=12$ & 27.70 & 0.20 & 41.80 & 0.20 \\
$n=0, k=16$ & 27.55 & 0.10 & \textbf{43.47} & 0.27 \\
$n=0, k=32$ & 27.57 & 0.27 & 41.93 & 0.47 \\
\addlinespace

$n=1, k=4$  & 27.13 & 0.13 & 41.73 & 0.33 \\
$n=1, k=8$  & \textbf{27.93} & 0.08 & 42.00 & 0.00 \\
$n=1, k=12$ & 27.58 & 0.18 & 42.33 & 0.73 \\
$n=1, k=16$ & 27.45 & 0.25 & 43.13 & 0.93 \\
$n=1, k=32$ & 27.60 & 0.20 & 41.87 & 0.53 \\
\addlinespace

$n=2, k=4$  & 27.18 & 0.72 & 41.80 & 0.60 \\
$n=2, k=8$  & 27.87 & 0.08 & 41.93 & 1.27 \\
$n=2, k=12$ & 27.78 & 0.33 & 41.80 & 0.80 \\
$n=2, k=16$ & 27.72 & 0.43 & 42.60 & 0.40 \\
$n=2, k=32$ & 27.50 & 0.20 & 41.73 & 1.27 \\
\addlinespace

$n=5, k=4$  & 27.18 & 0.57 & 41.80 & 0.00 \\
$n=5, k=8$  & 27.87 & 0.38 & 41.60 & 0.80 \\
$n=5, k=12$ & 27.80 & 0.30 & 41.93 & 0.93 \\
$n=5, k=16$ & 27.75 & 0.20 & 43.33 & 1.13 \\
$n=5, k=32$ & 27.63 & 0.62 & 42.07 & 0.33 \\
\addlinespace

$n=10, k=4$  & 27.20 & 0.25 & 42.00 & 0.60 \\
$n=10, k=8$  & 27.70 & 0.05 & 42.47 & 0.13 \\
$n=10, k=12$ & 27.77 & 0.33 & 42.53 & 0.93 \\
$n=10, k=16$ & 27.55 & 0.30 & 42.20 & 1.20 \\
$n=10, k=32$ & 27.45 & 0.30 & 41.93 & 0.67 \\
\addlinespace

$n=15, k=4$  & 27.38 & 0.18 & 41.53 & 0.53 \\
$n=15, k=8$  & 27.75 & 0.10 & 42.13 & 0.93 \\
$n=15, k=12$ & \textbf{27.93} & 0.27 & 42.27 & 0.53 \\
$n=15, k=16$ & 27.70 & 0.50 & 42.87 & 0.13 \\
$n=15, k=32$ & 27.57 & 0.18 & 41.87 & 2.07 \\

\end{longtable}
\endgroup

\FloatBarrier

\section{Runtime breakdown} \label{app:runtime}

\begin{table}[h]
\centering
\setlength{\tabcolsep}{8pt}
\caption{Wall-clock runtime per task averaged over three seeds. Subscript reports the maximum absolute deviation from the mean across seeds (in minutes).}
\renewcommand{\arraystretch}{1.05}
\begin{tabular}{lc}
\toprule
\textbf{Task} & \textbf{Avg runtime (min)} \\
\midrule
SST2     & 36.42\textsubscript{\textcolor{gray}{$\pm$0.43}} \\
SST5     & 44.00\textsubscript{\textcolor{gray}{$\pm$1.05}} \\
SUBJ     & 41.26\textsubscript{\textcolor{gray}{$\pm$1.69}} \\
CR       & 39.20\textsubscript{\textcolor{gray}{$\pm$2.08}} \\
MR       & 39.99\textsubscript{\textcolor{gray}{$\pm$1.30}} \\
NEWS     & 156.17\textsubscript{\textcolor{gray}{$\pm$6.89}} \\
TREC     & 9.98\textsubscript{\textcolor{gray}{$\pm$0.05}} \\
GSM8K    & 243.78\textsubscript{\textcolor{gray}{$\pm$2.96}} \\
MATH500  & 38.71\textsubscript{\textcolor{gray}{$\pm$0.60}} \\
MedQA    & 42.17\textsubscript{\textcolor{gray}{$\pm$0.60}} \\
DeepMath & 104.18\textsubscript{\textcolor{gray}{$\pm$2.12}} \\
\bottomrule
\end{tabular}
\label{tab:runtime}
\end{table}

\end{document}